\documentclass{article}

\usepackage{PRIMEarxiv}

\usepackage[utf8]{inputenc} % allow utf-8 input
\usepackage[T1]{fontenc}    % use 8-bit T1 fonts
\usepackage{hyperref}       % hyperlinks
\usepackage{url}            % simple URL typesetting
\usepackage{booktabs}       % professional-quality tables
\usepackage{amsfonts}       % blackboard math symbols
\usepackage{amsmath} 
\usepackage{nicefrac}       % compact symbols for 1/2, etc.
\usepackage{microtype}      % microtypography
\usepackage{lipsum}
\usepackage{graphicx}
\usepackage{microtype}      % microtypography
\usepackage{multirow}       % for multi-row cells in tables
\graphicspath{{media/}}     % organize your images and other figures under media/ folder

\title{MSMA: Multi-Scale Feature Fusion for Multi-Attribute 3D Face Reconstruction from Unconstrained Images}

\author{
  Danling Cao \\
  The Hong Kong Polytechnic University\\
     \And
  Hujun Yin \\
  The University of Manchester \\
}

\begin{document}
\maketitle

\begin{abstract}
Reconstruction 3D face from a single unconstrained image remains a challenging problem due to diverse conditions in unconstrained environments. Recently, learning-based methods have achieved notable results by effectively capturing complex facial structures and details across varying conditions. Consequently, many existing approaches employ projection-based losses between generated and input images to constrain model training. However, learning-based methods for 3D face reconstruction typically require substantial amounts of 3D facial data, which is difficult and costly to obtain. Consequently, to reduce reliance on labeled 3D face datasets, many existing approaches employ projection-based losses between generated and input images to constrain model training. Nonetheless, despite these advancements, existing approaches frequently struggle to capture detailed and multi-scale features under diverse facial attributes and conditions, leading to incomplete or less accurate reconstructions. In this paper, we propose a Multi-Scale Feature Fusion with Multi-Attribute (MSMA) framework for 3D face reconstruction from unconstrained images. Our method integrates multi-scale feature fusion with a focus on multi-attribute learning and leverages a large-kernel attention module to enhance the precision of feature extraction across scales, enabling accurate 3D facial parameter estimation from a single 2D image. Comprehensive experiments on the MICC Florence, Facewarehouse and custom-collect datasets demonstrate that our approach achieves results on par with current state-of-the-art methods, and in some instances, surpasses SOTA performance across challenging conditions.
\end{abstract}

\section{Introduction}
Reconstructing 3D faces from a single unconstrained 2D image is a fundamental yet challenging task in computer vision, with applications in augmented reality, virtual reality, and human-computer interaction. The 3D Morphable Model \cite{facemodel1999} laid the foundation for 3D face reconstruction by introducing a parameterized representation that encodes facial geometry and texture into a low-dimensional space. This statistical framework provided a structured and interpretable approach to modeling 3D facial attributes, enabling the prediction of facial shapes and textures from 2D images. The integration of deep learning into single-view 3D face reconstruction has marked a paradigm shift in the field. By leveraging convolutional neural networks, recent methods predict 3DMM coefficients directly from 2D image features, achieving significant advancements in reconstruction accuracy and adaptability to challenging imaging conditions. However, traditional CNN-based approaches typically rely on supervised learning pipelines requiring large-scale 3D datasets, such as meshes or point clouds \cite{dou2017end, tfmeshrender_2018_CVPR, Guo20193DFace, feng2021learning, jiang2023pointgs, zhao2023divide}. The collection and annotation of such high-quality 3D data remain resource-intensive, presenting barriers to scalability and practical deployment.

To mitigate this reliance on extensive 3D data, the advent of differentiable rendering techniques \cite{nvdiffrast2020, torch3d2020} has enabled weakly-supervised and self-supervised approaches. These methods integrate differentiable renderers into the learning framework, allowing for end-to-end optimization by constraining 3D reconstructions with 2D image-based losses, such as projections and landmarks. This significantly reduces the need for labeled 3D data while maintaining robust reconstruction fidelity across varying conditions \cite{tewari2018self, deep3d_deng, basak20223d, wood20223d}.

Despite these advancements, existing methods face notable challenges in balancing fine-grained detail capture with global structural consistency. 3D face reconstruction involves predicting multiple interdependent attributes—such as identity, expression, pose, texture, and illumination of which requires specialized features from different resolutions. Current methods often adopt a "one-size-fits-all" approach, where a single feature representation is used for all tasks. This can lead to suboptimal performance, as attributes like identity and depth depend on high-level structural features, while texture and illumination rely on fine-grained details.

To address these challenges, we propose the Multi-Scale Feature Fusion with Multi-Attribute (MSMA) framework, which leverages hierarchical features at multiple resolutions to disentangle and accurately model interdependent facial attributes. MSMA integrates multi-scale feature fusion with a novel task-specific learning mechanism, ensuring that each attribute is derived from its most relevant resolution. For instance, high-level features capture global facial geometry, mid-level features balance spatial and semantic information, and low-level features provide the local details necessary for fine-grained attributes. This design not only enhances reconstruction fidelity but also improves robustness to variations in pose, illumination, and occlusion.

Moreover, we incorporate a large-kernel attention mechanism to refine fused features by capturing long-range dependencies. This enhances the model's ability to maintain structural coherence and effectively disentangle complex facial attributes. The multi-attribute learning module further assigns task-specific regression heads for identity, expression, pose, and texture, ensuring precise reconstruction tailored to each attribute.

The key contributions of this work can be summarized as follows:
\begin{enumerate}
 \item Multi-Scale Feature Fusion with Multi-Attribute Learning. We propose MSMA, a novel framework that integrates multi-scale feature fusion with multi-attribute learning. By leveraging hierarchical features at different resolutions, MSMA achieves a balance between capturing fine-grained details and maintaining global consistency, addressing the challenges of 3D face reconstruction from unconstrained 2D images.
 \item Attribute-Specific Task Learning with Large-Kernel Attention. MSMA incorporates large-kernel attention mechanisms to refine fused features by capturing long-range dependencies. Combined with multi-scale fusion, the framework assigns task-specific regression heads to each attribute (e.g., identity, expression, pose, and texture), ensuring that the most relevant feature scale is utilized for accurate and robust attribute prediction.
 \item Comprehensive Evaluation. We conduct extensive experiments on multiple datasets, including MICC Florence, FaceWarehouse, MoFA-Test, AFLW2000-3D, and a custom-collected dataset. Through comparisons with ten state-of-the-art methods, our framework demonstrates competitive performance in both 3D face reconstruction and dense alignment tasks, showcasing improved accuracy, robustness, and generalizability across diverse real-world scenarios.
\end{enumerate}

\section{Related Work}
\label{sec:headings}
  \subsection{3D Face Morphable Model}
The 3D Morphable Model (3DMM), first introduced by \cite{facemodel1999}, has become a foundational framework in the field of 3D face reconstruction. It provides a statistical paradigm for representing facial shape and texture in a parametric form. Specifically, 3DMM \cite{survey} encodes the 3D geometry and texture of a face as a linear combination of deformation bases, allowing for the adjustment of parameters to model individual-specific facial features. By projecting the complex 3D structure of a face into a low-dimensional parameter space, 3DMM significantly reduces the computational burden associated with 3D reconstruction tasks. The widely adopted Basel Face Model (BFM) \cite{bfm2009} leverages principal component analysis (PCA) to derive basis vectors from facial shape and texture data, facilitating a parametric generation process. This formulation effectively transforms the challenging 2D-to-3D reconstruction problem into a more manageable parametric regression task while offering a clear mathematical foundation for optimizing model performance. Over the years, numerous enhanced versions of 3DMM have been developed to address its limitations and extend its applicability. Notable examples include FaceWarehouse \cite{facewarehouse2013}, FLAME \cite{flame2017}, the Surrey Face Model \cite{surrey2016}, and LSFM \cite{lsfm2018}. These advanced models have aimed to enhance the versatility, diversity, and robustness of the 3D Morphable Model.

\subsection{Single-View Learning-Based 3D Face Reconstruction}
Recent advancements in learning-based approaches have revolutionized 3D face reconstruction by enabling direct regression of 3D Morphable Model (3DMM) coefficients from a single 2D image. These methods have significantly improved computational efficiency and adaptability across diverse imaging conditions. Frameworks utilizing convolutional neural networks (CNNs) \cite{deep3d_deng, withoutsupervision2019, EMOCA2022, Li_2023_CVPR, 3DDFA-v3} and graph convolutional networks (GCNs) \cite{Lin_2020_CVPR, Gao_2020_CVPR_Workshops, Lee_2020_CVPR, Qiu_2021_CVPR} effectively capture the distribution of facial features, focusing primarily on refining 3D geometry. Embedding statistical parameters into differentiable rendering frameworks further enables robust end-to-end training and enhances geometric accuracy.

Many approaches utilize 3DMM to enhance facial geometry, a critical component of 3D face reconstruction \cite{Tran_2017_CVPR, withoutsupervision2019, Yi_2019_CVPR, MICA}. While some methods focus solely on geometry refinement, others aim for a holistic representation by addressing both geometric and texture-related challenges. However, these frameworks often introduce additional complexities that do not directly contribute to enhanced shape fidelity.

To enable accurate 3D face reconstruction, learning-based methods often integrate key submodels to disentangle facial geometry, pose, and illumination parameters. This section outlines the face model, camera model, and illumination model components incorporated in our framework.

\textbf{Face Model.}
In the 3D Morphable Model (3DMM), a 3D face is parameterized using shape and texture bases. The shape $\mathbf{S}$ and texture $\mathbf{T}$ of a 3D face are represented as:

\begin{equation}
    \begin{aligned}&\mathbf{S}(\boldsymbol{\alpha},\boldsymbol{\beta}_{exp})=\overline{\mathbf{S}}+\mathbf{A}_{id}\boldsymbol{\alpha}_{id}+\mathbf{A}_{exp}\boldsymbol{\beta}_{exp}\\&\mathbf{T}(\boldsymbol{\gamma}_{tex})=\overline{\mathbf{T}}+\mathbf{A}_{tex}\boldsymbol{\gamma}_{tex}\end{aligned}
\end{equation}

where $\overline{\mathbf{S}}$ and $\overline{\mathbf{T}}$ are the mean shape and texture, $\mathbf{A}_{id}$, $\mathbf{A}_{exp}$, and $\mathbf{A}_{tex}$ are the shape bases, expression bases, and texture bases, respectively. $\boldsymbol{\alpha}_{id}\in\mathbb{R}^{80}$, $\boldsymbol{\beta}_{exp}\in\mathbb{R}^{64}$, and $\boldsymbol{\gamma}_{tex}\in\mathbb{R}^{80}$ are the coefficients of the shape, expression, and texture, respectively. We employ the widely-used 2009 Basel Face Model(BFM) as the underlying 3DMM for $\overline{\mathbf{S}}$, $\overline{\mathbf{T}}$, $\mathbf{A}_{id}$, and $\mathbf{A}_{tex}$, excluding the ear and neck regions. For the expression model $\mathbf{A}_{exp}$, we adopt the bases built from the FaceWarehouse dataset, utilizing delta blendshapes to represent facial expressions.

\textbf{Camera Model.}
We adopt a weak perspective projection model to transform 3D face geometry into 2D image space. The projected geometry $\mathbf{V_{2d}}$ is calculated as:

\begin{equation}
    \mathbf{V_{2d}} = \mathbf{P_r} * (\mathbf{R*S+t})
\end{equation}

where $\mathbf{P_r}$ is the projection matrix, and $\mathbf{S}$ is the pose-independent face shape. $\mathbf{R}$ and $\mathbf{t}$ are the 3D face rotation matrix and translation matrix, respectively.We implement a perspective camera model with an empirically determined focal length for the 3D-to-2D projection geometry. Therefore, the 3D face pose $\boldsymbol{\theta}_{pos}$ is parameterized by an Euler rotation $\mathbf{R} \in \mathbf{SO(3)}$ and translation $t \in \mathbf{R}^3$.

\textbf{Illumination Model.}
To achieve realistic face rendering, we approximate scene illumination using Spherical Harmonics(SH) following \cite{SH2001}. Similar to \cite{deep3d_deng}, we choose the first three bands of SH basis functions, and using the face texture and surface normal as input, then the shaded texture $\mathbf{T_{sh}}$ is computed as follows:

\begin{equation}
    \mathbf{T_{sh}} = \mathbf{T}(\boldsymbol{\gamma}_{tex}) \odot \sum_{k=1}^9 \boldsymbol{\delta}_{lig} \Psi_k(\mathbf{n})
\end{equation}

where \( \odot \) denotes the Hadamard product, $\mathbf{n}$ represents the surface normals of the 3D face model, \( \Psi_k : \mathbb{R}^3 \to \mathbb{R} \) is the SH basis function, and $\boldsymbol{\delta}_{lig} \in \mathbf{R}^9$ is the corresponding SH parameters.

  \subsection{Multi-Scale Feature Fusion}
  Multi-scale feature fusion is a widely adopted strategy in image-based vision tasks due to its ability to integrate information across resolutions, capturing both global and local details essential for downstream predictions. Early approaches primarily focused on combining features from encoding and decoding stages, such as adding downsampled features from the encoder with upsampled features in the decoder to refine spatial details in human pose estimation \cite{PoseEsti}. Recent advancements have introduced more sophisticated fusion strategies tailored for diverse tasks. For instance, HRNet \cite{HRNet} maintains parallel multi-resolution feature streams and fuses them iteratively to enhance spatial consistency for tasks like semantic segmentation and object detection. In medical image segmentation, channel-wise concatenation of features from different scales followed by 1×1 convolutions has been shown to effectively exploit multi-scale contextual information while preserving computational efficiency. Extending these principles, our framework incorporates multi-scale fusion to align hierarchical features across resolutions, enabling a more precise disentanglement of interdependent facial attributes crucial for face reconstruction.

  \subsection{Attention Mechanism in Visual Tasks}
  Attention mechanisms were initially introduced in natural language processing (NLP) to enhance feature extraction by focusing on the most relevant parts of the input sequence \cite{attention2017Vaswani}. Their success in NLP inspired their adaptation to computer vision tasks, where attention mechanisms enable models to prioritize critical image regions, improving performance in image recognition, object detection, and semantic segmentation \cite{atteninvision}.
  
  In CNN-based architectures, self-attention mechanisms have shown particular effectiveness in later layers, where they capture fine-grained details and long-range dependencies \cite{2023largeseparablekernelattention}. Larger convolutional kernels naturally expand the receptive field, offering improved spatial context modeling. For example, LRNet \cite{LRNet2019} utilized 7×7 kernels to enhance relational modeling, demonstrating the benefits of larger kernels for local feature relationships. However, large kernels are computationally expensive. To address this, Guo et al. \cite{guo2022visualattentionnetwork} introduced cascaded depthwise convolutions and dilated convolutions to efficiently emulate large receptive fields without incurring excessive computational costs.
  
  In particular,  In CNN-based architectures, self-attention mechanisms, especially in later layers, significantly improve performance by capturing fine-grained details and global dependencies \cite{2023largeseparablekernelattention}. Larger convolutional kernels naturally expand the receptive field, enabling broader spatial context modeling. For instance, LRNet \cite{LRNet2019} used 7×7 kernels to enhance relational modeling. To address the high computational cost of large kernels, \cite{guo2022visualattentionnetwork} proposed cascaded depthwise convolutions and dilated convolutions to emulate large receptive fields efficiently. Recently, Multi-Scale Large Kernel Attention (MLKA) \cite{mlka2024} was introduced, grouping channels and applying progressively larger kernels to model multi-scale dependencies. This makes MLKA well-suited for hierarchical tasks like our multi-scale fusion and multi-attribute reconstruction, enabling both local detail capture and global coherence.

Recently, MLKA \cite{mlka2024} enhanced attention mechanisms by grouping feature channels and applying progressively larger kernels. In our framework, it is applied after fusion to refine multi-resolution features, facilitating the subsequent multi-attribute reconstruction.

 \begin{figure}[t]
      \centering
      \includegraphics[width=1.0\textwidth]{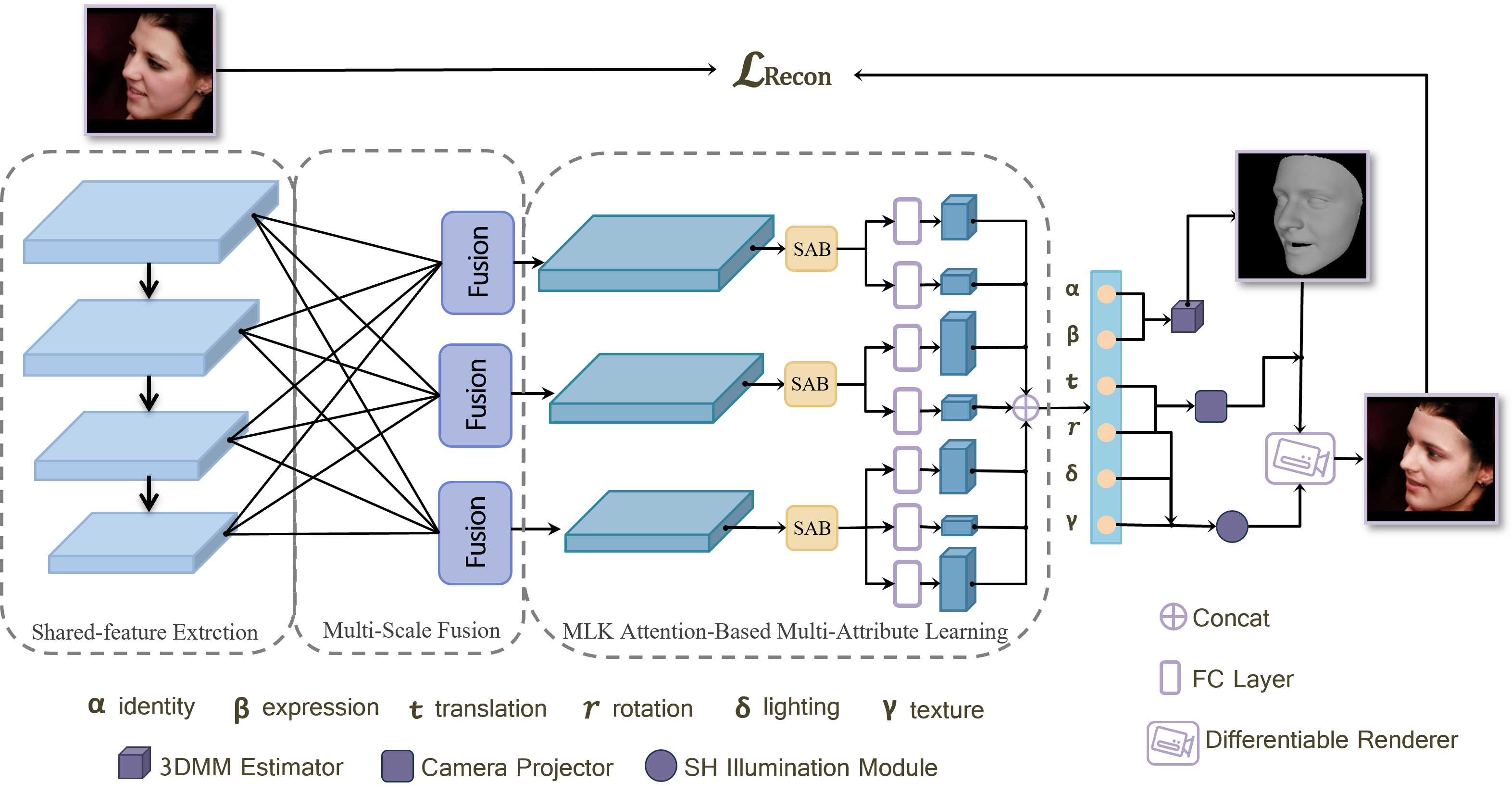}
      \caption[Overview of the Proposed Multi-Scale Feature Fusion with Multi-Attribute
(MSMA) framework]{\textbf{Overview of our method.}
    The MSMA framework first extracts multi-scale features from input images to address the challenges of capturing fine-grained facial details and maintaining global consistency. It then leverages two key components—Multi-Scale Fusion (MSF) and the MLK Attention-Based Multi-Attribute (MAMA) module—to refine feature representations and ensure accurate reconstruction of diverse facial attributes under unconstrained conditions. Finally, a differentiable renderer enforces alignment between the input image and the reconstructed output.}
      \label{fig:msma_overview}
\end{figure} 
\section{Method}
  \label{sec:Method}
  \subsection{Overview}
Building on the multi-scale feature extraction capability of the ResNet \cite{resnet2016} architecture, we propose the Multi-Scale Feature Fusion with Multi-Attribute (MSMA) framework, which is designed to address the challenges of capturing fine-grained facial shape and texture details while maintaining global structural consistency under unconstrained imaging conditions. The architecture of the MSMA framework is illustrated in Figure  \ref{fig:msma_overview},  consists of two main  components: Multi-Scale Fusion (MSF) and the MLK Attention-Based Multi-Attribute (MAMA) module. 

A key limitation of existing 3D face reconstruction methods is their inability to exploit hierarchical features effectively for different facial attributes, leading to inaccuracies in reconstructing interdependent properties like identity, texture, pose, and illumination. To address this, the MSF module extracts, aligns, and fuses multi-resolution features, combining global structural information with local fine-grained details. Low-resolution features focus on attributes requiring global coherence, such as identity and expression, while mid-resolution features enhance attributes like illumination and texture, which rely on detailed spatial information. This adaptive feature utilization improves both robustness and precision.

However, refining these fused features to capture long-range dependencies and local variations remains challenging. The MAMA module tackles this by integrating a Large Kernel Attention (LKA) mechanism to model global context and refine feature representations. Additionally, a multi-attribute learning strategy assigns task-specific regression heads to predict parameters. 
 
  \subsection{Multi-scale Feature Fusion (MSF) Module}
\begin{figure}[h]
  \centering
  \includegraphics[width=0.9\textwidth]{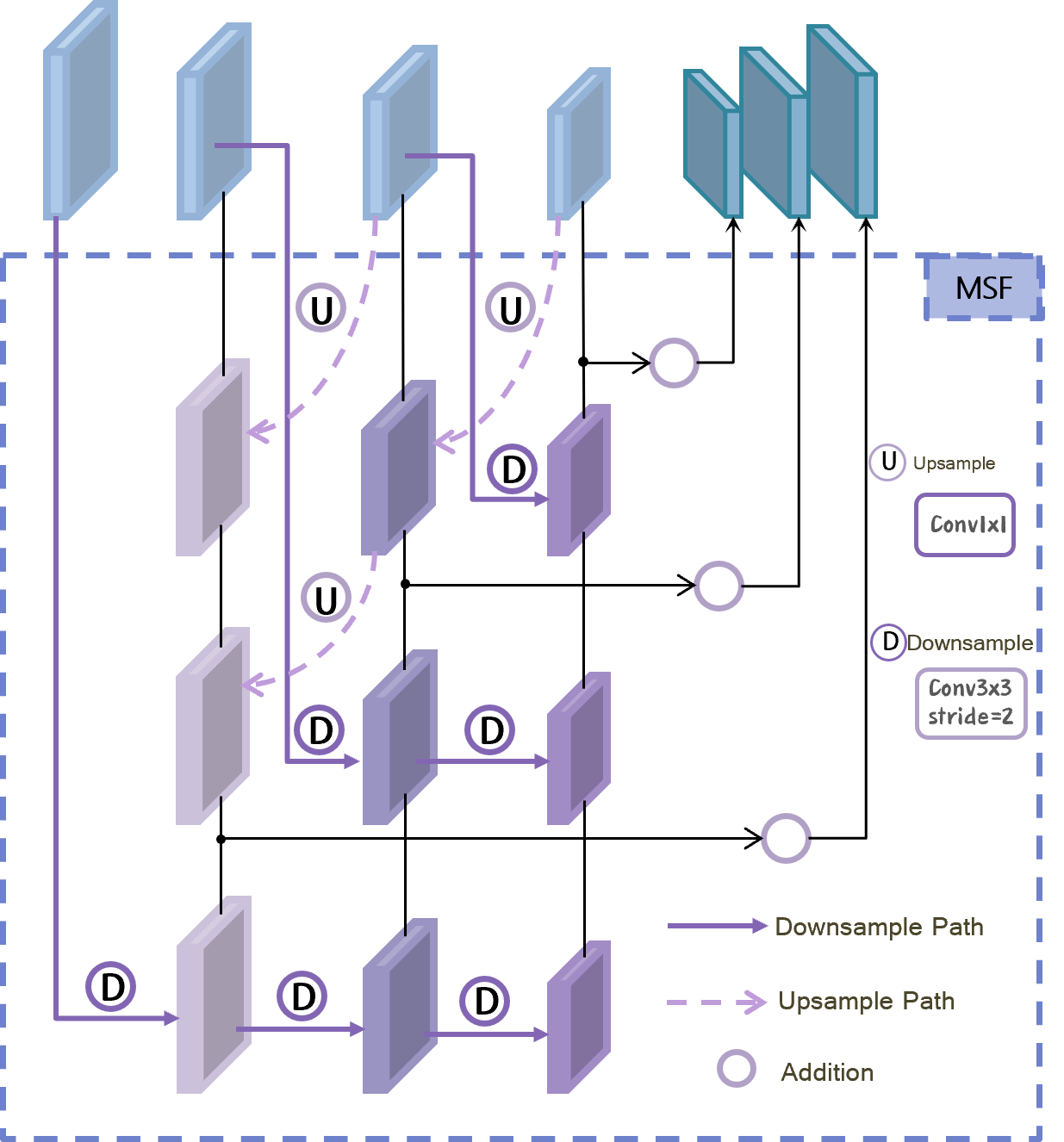}
  \caption[Overview of multi-scale feature fusion process]{\textbf{Multi-Scale Fusion Module (MSF Module).}
The module aligns and integrates hierarchical feature maps from a shared backbone network, combining features from multiple resolutions for enriched multi-scale representations. Inspired by HRNet \cite{HRNet}, it uses upsampling and downsampling to align feature sizes and channels, followed by element-wise addition for fusion.}
  \label{fig:Multi_fusion}
\end{figure}

The Multi-Scale Fusion (MSF) module is designed to align and integrate hierarchical feature maps extracted from a shared backbone network, as illustrated in \ref{fig:Multi_fusion}. By effectively combining features from multiple resolutions, the MSF module ensures enriched multi-scale representations, which are crucial for modeling intricate details and supporting subsequent multi-attribute reconstruction. The module retains the last three intermediate features, with fusion feature sizes of $28 \times 28 \times 512$, $14 \times 14 \times 1024$, and $7 \times 7 \times 2048$, ensuring a compact and expressive representation. Inspired by HRNet \cite{HRNet}, the MSF module employs a combination of upsampling and downsampling operations to align feature sizes and channels across resolutions, followed by element-wise addition for feature fusion.

Firstly, hierarchical feature maps are extracted from the backbone network, denoted as $\mathbf{F}_1$, $\mathbf{F}_2$, $\mathbf{F}_3$, $\mathbf{F}_4$. Here, $\mathbf{F}_1$ represents high-resolution features capturing fine spatial details extracted from the backbone's initial stage, while $\mathbf{F}_4$ encodes low-resolution features with global semantic information from the final stage. Similarly, $\mathbf{F}_2$ and $\mathbf{F}_3$ correspond to intermediate stages, encoding a balance of spatial and semantic features. To fuse the features into a target map $\mathbf{F}_i$, all other feature maps $\mathbf{F}_j$ ($j \neq i$) are aligned to match the spatial size and channel dimensions of $\mathbf{F}_i$, followed by element-wise addition.

Mathematically, for a target feature map $\mathbf{F}_i$, the fused feature map $\mathbf{F}_{fused,i}$ is computed as:

\begin{equation}
\mathbf{F}_{\text{fused}, i} = \sum_{j=1}^{4} \text{Align}(\mathbf{F}_j, \text{size}(\mathbf{F}_i), \text{channels}(\mathbf{F}_i)),
\end{equation}

where $\text{Align}(\cdot)$ represents either upsampling or downsampling to match the size and channel dimensions of $\mathbf{F}_i$.

The alignment process for multi-scale feature fusion can be mathematically formulated as follows:
For $\mathbf{F}_j$, where $j > i$ (requiring upsampling to match $\mathbf{F}_i$):

\begin{equation}
\text{Align}(\mathbf{F}_j) = \text{Interpolate}(\mathbf{F}_j, \text{size}(\mathbf{F}_i)),
\end{equation}

and for $\mathbf{F}_j$, where $j < i$ (requiring downsampling to match $\mathbf{F}_i$), the alignment is performed iteratively using $3 \times 3$ convolutions with a stride of $2$:

\begin{equation}
\text{Align}(\mathbf{F}_j) = \text{Conv}_{3 \times 3}^k(\mathbf{F}_j),
\end{equation}

After alignment, the fused feature map $\mathbf{F}_i$ is computed by summing all aligned feature maps, followed by a non-linear activation:

\begin{equation}
\mathbf{F}_{\text{fused}, i} = \text{ReLU}\left(\mathbf{F}_i + \sum_{j=1, j \neq i}^{4} \text{Align}(\mathbf{F}_j)\right).
\end{equation}

Finally, the MSF module retains the fused feature maps $\mathbf{F}_{fused,2}$, $\mathbf{F}_{fused,3}$, and $\mathbf{F}_{fused,4}$ as the intermediate features for subsequent modules. Among these, $\mathbf{F}_{fused,2}$preserves a balance between spatial and semantic information, $\mathbf{F}_{fused,3}$ emphasises mid-level semantics, and $\mathbf{F}_{fused,4}$ captures the global structural information of the input. Together, these fused features enable the MSF module to effectively integrate multi-scale information, enhancing the model’s capability to capture both global and local details while ensuring robust and precise attribute predictions in downstream tasks.
  \begin{figure}[t]
      \centering
      \includegraphics[width=1.0\textwidth, height=0.5\textheight]{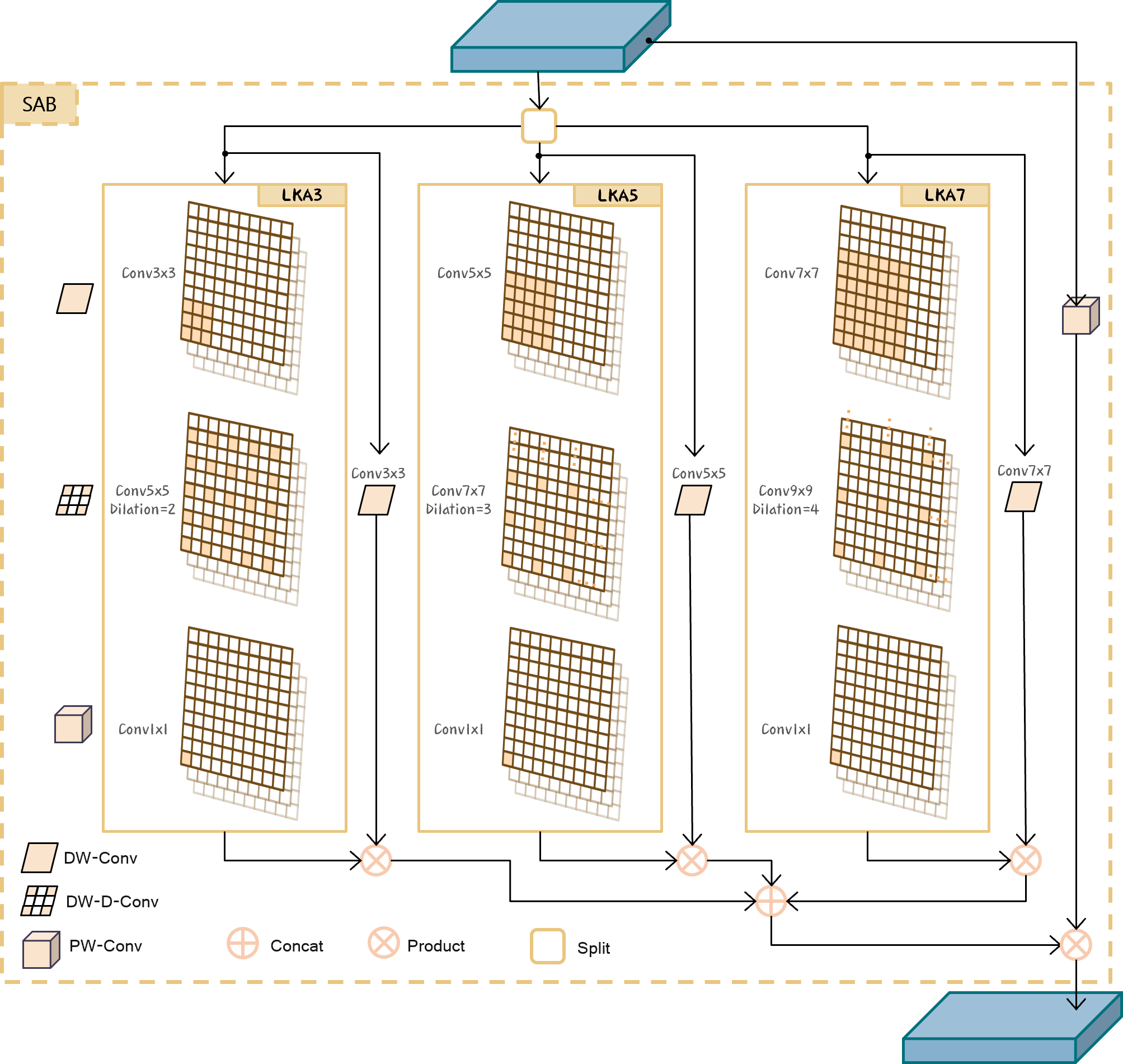}
      \caption[Overview of Self-Attention Block]{\textbf{Self-Attention Block with MLKA.}
    The PAFB integrates multi-scale features by combining high-resolution feature maps with upsampled low-resolution maps. This process involves spatial and channel attention mechanisms, enhancing the feature representation by emphasising important spatial locations and channels. The refined feature maps are then fused through element-wise operations for further processing.}
      \label{fig:SAB}
    \end{figure}  
  \subsection{MLK Attention-Based Multi-Attribute (MAMA) Module}
The MLK Attention-Based Multi-Attribute (MAMA) module is designed to disentangle and predict various facial attributes from the fused multi-scale features generated by the MSF module. This module ensures that hierarchical features extracted at different resolutions are effectively utilised for specific reconstruction tasks. By employing task-specific regression heads, MAMA achieves attribute-specific learning while preserving the interdependence of facial attributes.

    \textbf{Large Kernel Attention.}
After generating three resolution-specific feature maps through the multi-scale fusion process, the next step focuses on extracting multi-scale dependencies, extending the receptive field of CNN operations, and preserving the hierarchical structure of the fused representations. To achieve this, we incorporate the Multi-Scale Large Kernel Attention (MLKA) mechanism \cite{mlka2024}, initially developed for image super-resolution, as the self-attention block within the MAMA module. The MLKA module, illustrated in Figure \ref{fig:SAB}.

The MLKA module builds on depthwise separable convolutions, a key design choice to computationally efficient while expanding the receptive field. By decoupling spatial and channel-wise operations, depthwise separable convolutions capture extensive contextual information across scales without introducing a significant computational burden.

The Self-Attention Block (SAB) processes each input feature map through three parallel branches, each corresponding to kernel sizes of 3 x 3, 5 x 5, and 7 x 7, larger convolution kernels capture a larger receptive field. Larger kernels capture broader spatial dependencies, enhancing the network's ability to model long-range contextual relationships. Each branch is composed of three serial stages: a Depthwise Convolution stage, a Dilated Convolution stage, and a Pointwise Convolution stage. These stages are designed to: Apply spatial filtering to individual channels through depthwise convolutions, expand the receptive fields using dilated convolutions, with progressively increasing dilation rates for each kernel size (2 ,3,and 4), and refine features and reduce dimensionality through pointwise convolutions. The input feature map, denoted as $\mathbf{F}_{in}$, is evenly split into three segments, $\mathbf{F}_{3}$, $\mathbf{F}_{5}$, and $\mathbf{F}_{7}$, which serve as inputs to the three branches. The intermediate processing steps for each branch are formulated as:

\begin{equation}
\begin{aligned}
\mathbf{D}_{k} &= \text{DWConv}_{k}(\mathbf{F}_{k}) \ast \text{DilatedConv}_{k}(\mathbf{F}_{k}), \quad k \in \{3, 5, 7\}, \\
\mathbf{P}_{k} &= \text{PointwiseConv}_{1 \times 1}(\mathbf{D}_{k})
\end{aligned}
\end{equation}

The outputs from all three branches are concatenated along the channel dimension and fused via element-wise multiplication with their corresponding depthwise features. This process ensures that each spatial scale contributes proportionally to the final feature representation: 

\begin{equation}
\mathbf{F}_{\text{LKA}} = \text{Concat}(\text{LKA}_i(\mathbf{F}_i) \odot \mathbf{X}_i(\mathbf{F}_i)), \quad i \in \{3, 5, 7\}
\end{equation}

where $\text{LKA}_i$ denotes the large kernel attention block for kernel size \text{i}, $X_{i}$ represents the depthwise convolution kernel with size \text{i} x \text{i}, and $\odot$ is the element-wise multiplication.

Finally, the refined feature map is passed through a projection layer and combined with the input via a residual connection to preserve the original spatial information: 

\begin{equation}
\mathbf{F}_{\text{out}} = \mathbf{F}_{\text{in}} + \text{Proj}(\mathbf{F}_{\text{LKA}}) \cdot \text{Scale}
\end{equation}

where \text{Proj} represents a 1×1 convolutional layer for channel-wise refinement, and Scale is a learnable parameter.

    \textbf{Multi-Attribute Learning.}
As depicted in the latter half of Figure \ref{fig:msma_overview}, the multi-scale feature maps ($\mathbf{F}_1$, $\mathbf{F}_2$, $\mathbf{F}_3$) produced after multi-scale fusion and SAB are strategically utilized for attribute-specific regression tasks. High-resolution features ($\mathbf{F}_1$), which capture localized and fine-grained spatial details, are allocated to attributes such as lighting ($\delta$) and translation ($t$), as these tasks require high spatial precision. Mid-resolution features ($\mathbf{F}_2$), balancing semantic richness and spatial detail, are particularly suited for predicting pose ($r$) and texture ($\gamma$), where an equilibrium of global and local information is critical. Conversely, low-resolution features ($\mathbf{F}_3$) encode global structural information, making them ideal for attributes such as identity ($\alpha$) and expression ($\beta$), which demand coherence across the entire face.
The regression for each attribute is formulated as: 

\begin{equation}
\begin{aligned}
\alpha &= \text{FC}_{\alpha}(\mathbf{F}_3), & \beta &= \text{FC}_{\beta}(\mathbf{F}_3), \\
\gamma &= \text{FC}_{\gamma}(\mathbf{F}_2), & r &= \text{FC}_{r}(\mathbf{F}_2), \\
\delta &= \text{FC}_{\delta}(\mathbf{F}_1), & t &= \text{FC}_{t}(\mathbf{F}_1)
\end{aligned}
\end{equation}

where $\text{FC}_i$ denotes the fully connected layers tailored for each specific attribute regression task. To generalise the process, the prediction for an attribute can be expressed as:

\begin{equation}
\mathbf{A}_i = \text{RegressionHead}_i(\mathbf{F}_i), \quad i \in {\alpha, \beta, t, r, \delta, \gamma} 
\end{equation} 

where $\mathbf{A}_i$ represents the predicted coefficient for the attribute $i$, and $\text{RegressionHead}_i$ is the task-specific regression module that maps the feature $\mathbf{F}_i$ to the target parameter space.

To preserve interdependence among attributes while ensuring task-specific optimization, the predicted coefficients are aggregated through feature concatenation:

\begin{equation} \mathbf{A}{\text{concat}} = \text{Concat}(\mathbf{A}_{\alpha}, \mathbf{A}_{\beta}, \mathbf{A}_{t}, \mathbf{A}{r}, \mathbf{A}_{\delta}, \mathbf{A}_{\gamma}) 
\end{equation}

The concatenated coefficients, encompassing the predictions for all facial attributes, are passed to a differentiable renderer to reconstruct detailed 3D facial geometry and texture. This end-to-end framework seamlessly integrates feature extraction, attribute regression, and rendering, enabling joint optimization across these interconnected processes. 

  \section{Weakly Supervised Training}
    \label{sec:Weakly Supervised Training}
 The training objective for reconstructing a 3D face from an input image $\mathbf{I}$ is designed to combine multiple loss terms, each addressing specific aspects of the reconstruction process. This approach mitigates the reliance on extensive labeled 3D data by leveraging weakly supervised signals. The core idea is to project the predicted 3D face model back into 2D space and compare it with the original image. These loss terms collectively guide the model to balance geometric accuracy, texture detail, and alignment precision. The total loss \( \mathcal{L} \) is defined as follows:
 
\begin{equation}
\begin{aligned}
    \mathcal{L} = \lambda_{\text{pho}} \mathcal{L}_{\text{pho}} + \lambda_{\text{per}} \mathcal{L}_{\text{per}} + \lambda_{\text{lmk}} \mathcal{L}_{\text{lmk}} \\
    + \lambda_{\text{3dmm}} \mathcal{L}_{\text{3dmm}} + \lambda_{\text{refl}} \mathcal{L}_{\text{refl}}
\end{aligned}
\end{equation}

where the $\mathcal{L}_{\text{pho}}$ is the photometric loss, $\mathcal{L}_{\text{per}}$ is the perceptual loss, $\mathcal{L}_{\text{lmk}}$ is the landmark reprojection loss, $\mathcal{L}_{\text{3dmm}}$ and $\mathcal{L}_{\text{refl}}$ is the 3dmm coefficients regularization loss, and $\mathcal{L}_{\text{refl}}$ is the reflectance loss. 
  \subsection{Photometric Loss}
  Inspired by \cite{deep3d_deng}, we utilize the differentiable renderer \cite{nvdiffrast2020} to obtain the rendered image $mathbf{I}^{r}$. We use the $l_{2}$ loss to compute the photometric discrepancy between the input image $\mathbf{I}$ and the rendered image $\mathbf{I^{r}}$. The photometric loss is defined as:
  
\begin{equation}
    \mathcal{L}_{\text{pho}} = \frac{\sum_{i \in \mathcal{M}} A_i \cdot \left\| \mathbf{I} - \mathbf{I}^r \right\|_2}{\sum_{i \in \mathcal{M}} A_i},
\end{equation}

where i represents the pixel index, and M denotes the reprojected face region generated by the differentiable renderer. A is a face mask with a value of 1 in the face skin region and a value of 0 elsewhere obtained by an existing face segmentation method \cite{saito2016real}, which is capable of reducing errors caused by occlusion, such as eyeglasses.
  \subsection{Perceptual Loss}
  We employ the ArcFace model \cite{arcface} for feature embedding to compute the perceptual loss. The perceptual loss is defined as:
  
\begin{equation}
    \mathcal{L}_{\text{per}} = 1 - \frac{\Gamma(\mathbf{I}) \cdot \Gamma(\mathbf{I}^r)}{\|\Gamma(\mathbf{I})\|_2 \cdot \|\Gamma(\mathbf{I}^r)\|_2},
\end{equation}

where $\Gamma(\cdot)$ denotes the feature embedding obtained from the ArcFace model, $\mathbf{I}$ is the input image, and $\mathbf{I}^r$ is the rendered image. The loss measures the cosine similarity between the feature embeddings of the input and rendered images.
  \subsection{Landmark Loss}
  We utilize 68 facial landmarks, automatically detected from the input image, as weak supervision. For each training image, the face alignment network [] detects the 2D positions of these landmarks, denoted as $\{\mathbf{p}_n\}$. To calculate the reprojection loss during training, we project the 3D landmarks of the reconstructed face shapes onto the 2D image plane to obtain the corresponding 2D landmarks, denoted as $\{\mathbf{p}'_n\}$. The reprojection loss is calculated as follows:
  
\begin{equation}
    \mathcal{L}_{\text{lmk}} = \frac{1}{N} \sum_{n=1}^{N} \omega_n \left\| \mathbf{p}_n - \mathbf{p}'_n \right\|_2^2,
\end{equation}

where \( N \) is the number of facial landmarks (68), and \( \omega_n \) is the weight assigned to the \( n \)-th landmark, which we set to 20 for the inner mouth and 1 for the others \cite{deep3d_deng}. 
  \subsection{Regularization Loss}
In order to preserve the integrity of facial shape and texture, we employ a regularization loss on the estimated 3DMM coefficients to enforce a prior distribution towards the mean face. The coefficient loss is defined as:

\begin{equation}
    \mathcal{L}_{\text{3DMM}}(\mathbf{x}) = \lambda_{\alpha} \left\| \alpha_{id} \right\|_2^2 + \lambda_{\beta} \left\| \beta_{exp} \right\|_2^2 + \lambda_{\gamma} \left\| \gamma_{tex} \right\|_2^2
\end{equation}

Following \cite{deep3d_deng}, we set the weights \( \lambda_{\alpha} = 1.0 \), \( \lambda_{\beta} = 0.8 \), and \( \lambda_{\gamma} = 1.7e-2 \), and through the reflectance loss to favor a constant skin albedo similar to \cite{tewari2018self} . Which is formulated as:

\begin{equation}
    \mathcal{L}_{\text{refl}} = \frac{\sum (\mathbf{M} \odot (\mathbf{T} - \overline{\mathbf{T}}))^2}{\sum \mathbf{M}}
\end{equation}

where \(\overline{\mathbf{T}} = \frac{\sum (\mathbf{M} \odot \mathbf{T})}{\sum \mathbf{M}}\) is the mean texture within the masked region, \(\mathbf{M}\) is a binary mask indicating the face region, and \(\odot\) denotes element-wise multiplication. 

The overall regularization loss combines these two components:

\begin{equation}
    \mathcal{L}_{\text{reg}} = \lambda_{\text{3dmm}} \mathcal{L}_{\text{3dmm}} + \lambda_{\text{refl}} \mathcal{L}_{\text{refl}}
\end{equation}

where $\lambda_{\text{3dmm}}=3e-4$ and $\lambda_{\text{refl}} = 5.0$ are the balance weights between the 3DMM coefficients loss and the reflectance loss.

  \section{Experiments and Results}
  \label{sec:Experiments and Results}
  \subsection{Implementation Details}
    We implement our method using on PyTorch framework \cite{paszke2019pytorch} and employed the differentiable renderer from Nvdiffrast \cite{nvdiffrast2020}. Training was conducted with the Adam optimizer \cite{adam} at an initial learning rate of 0.0004 and a batch size of 64. To facilitate stable convergence, we decayed the learning rate by a factor of 0.5 every 10 epochs. Input images were color face images resized to 224 × 224 × 3. All experiments were executed on an NVIDIA Tesla V100 GPU.

    \subsection{Datasets}
  To train our model, we used two publicly available datasets: LFW \cite{lwf} and VGGFace2 \cite{vggface2}. The Labeled Faces in the Wild (LFW) dataset comprises  over 13,000 face images collected from diverse online sources, while the VGGFace2 dataset includes approximately 3.31 million images from 9,131 unique subjects. For training, we filtered and preprocessed these datasets by randomly selecting a subset of images, on which we applied the FAN \cite{fan2017point} to extract 68 2D facial landmarks as ground truth. Images that did not meet detection criteria or contained multiple detected bounding boxes were discarded, resulting in a curated training set of approximately 50,000 single face images. To assess the 3D face geometric reconstruction performance of our model, we utilized the MICC Florence \cite{Florence} and FaceWarehouse \cite{facewarehouse2013} datasets.

  \subsection{Qualitative Results}
To qualitatively evaluate the performance of our proposed MSMA framework, we compared it with previous ten methods, including Sela \textit{et al.} \cite{sela2017unrestricted}, L.Tran \cite{Tran_2017_CVPR}, Tewari \textit{et al.} \cite{tewari2018self}, MoFA \cite{MoFA_Tewari_2017_ICCV}, Tran \textit{et al.} \cite{tran2018nonlinear}, Genova \textit{et al.} \cite{tfmeshrender_2018_CVPR}, and Gecer \textit{et al.} \cite{Gecer_2019_CVPR}, PRNet \cite{feng2018prn}, 3DDFA-V2 \cite{3DDFAv2}, and Yu \textit{et al.} \cite{deep3d_deng}. The evaluation was conducted on the MoFA test dataset \cite{MoFA_Tewari_2017_ICCV}, AFLW2000-3D dataset \cite{ALFW-3D}, and a custom-collected dataset under various challenging conditions, such as occlusions (e.g., bangs and sunglasses), different facial expressions, and significant head poses. 
\begin{figure}[t]
  \centering
  \includegraphics[width=1.0\textwidth, height=0.6\textheight]{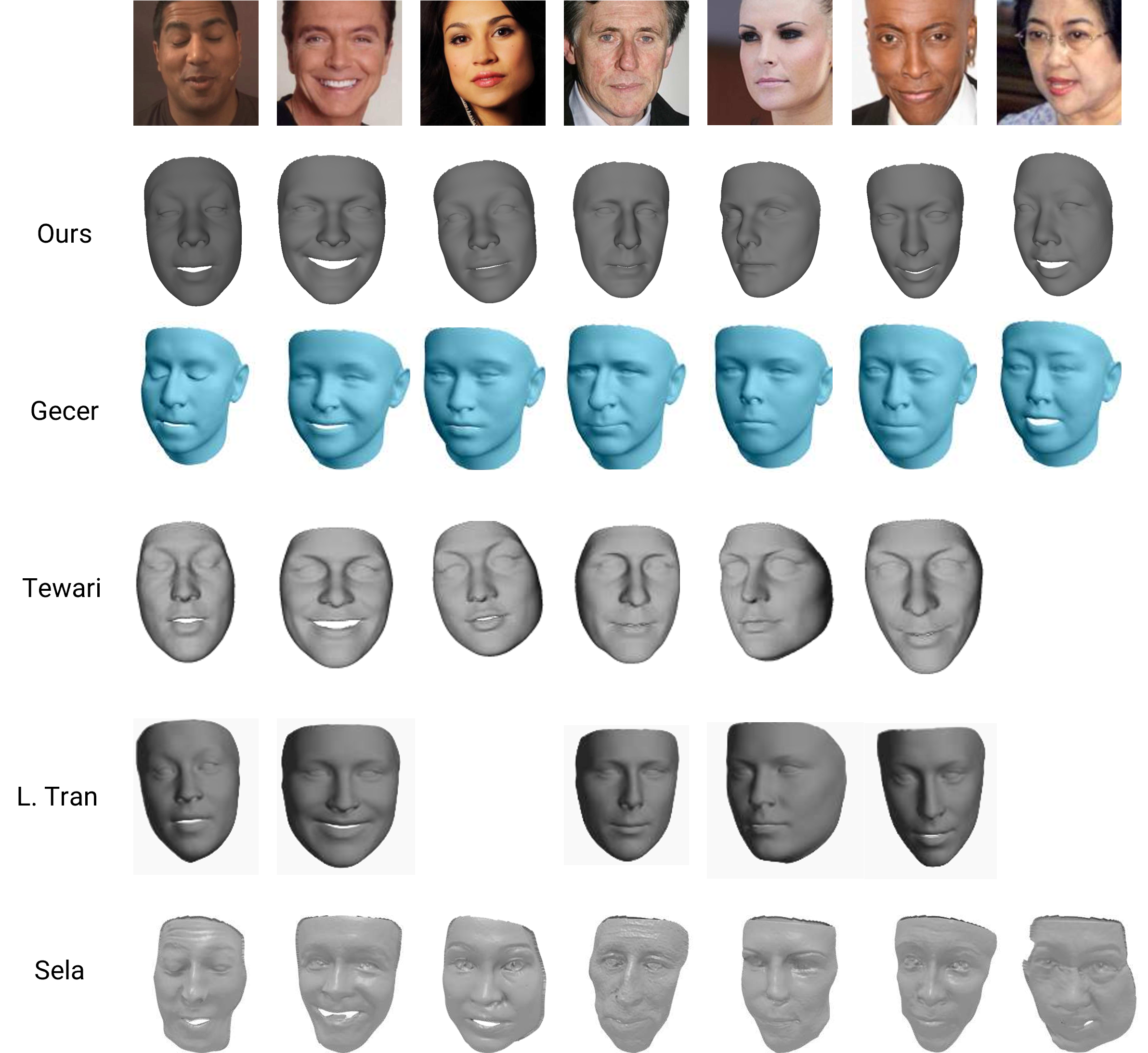}
  \caption[Qualitative comparison of rendered texture on MoFA-Test dataset]{Qualitative comparison of rendered texture results on the MoFA-Test dataset. Results from previous methods are taken from \cite{tfmeshrender_2018_CVPR, Gecer_2019_CVPR}, including Sela \textit{et al.} \cite{sela2017unrestricted}, Tewari \textit{et al.} \cite{tewari2018self}, MoFA \cite{MoFA_Tewari_2017_ICCV}, Tran \textit{et al.} \cite{Tran_2017_CVPR}, Genova \textit{et al.} \cite{tfmeshrender_2018_CVPR}, and Gecer \textit{et al.} \cite{Gecer_2019_CVPR}.}
  \label{fig:mofa_shape_compar}
\end{figure} 
\begin{figure}[t]
  \centering
  \includegraphics[width=1.0\textwidth, height=0.6\textheight]{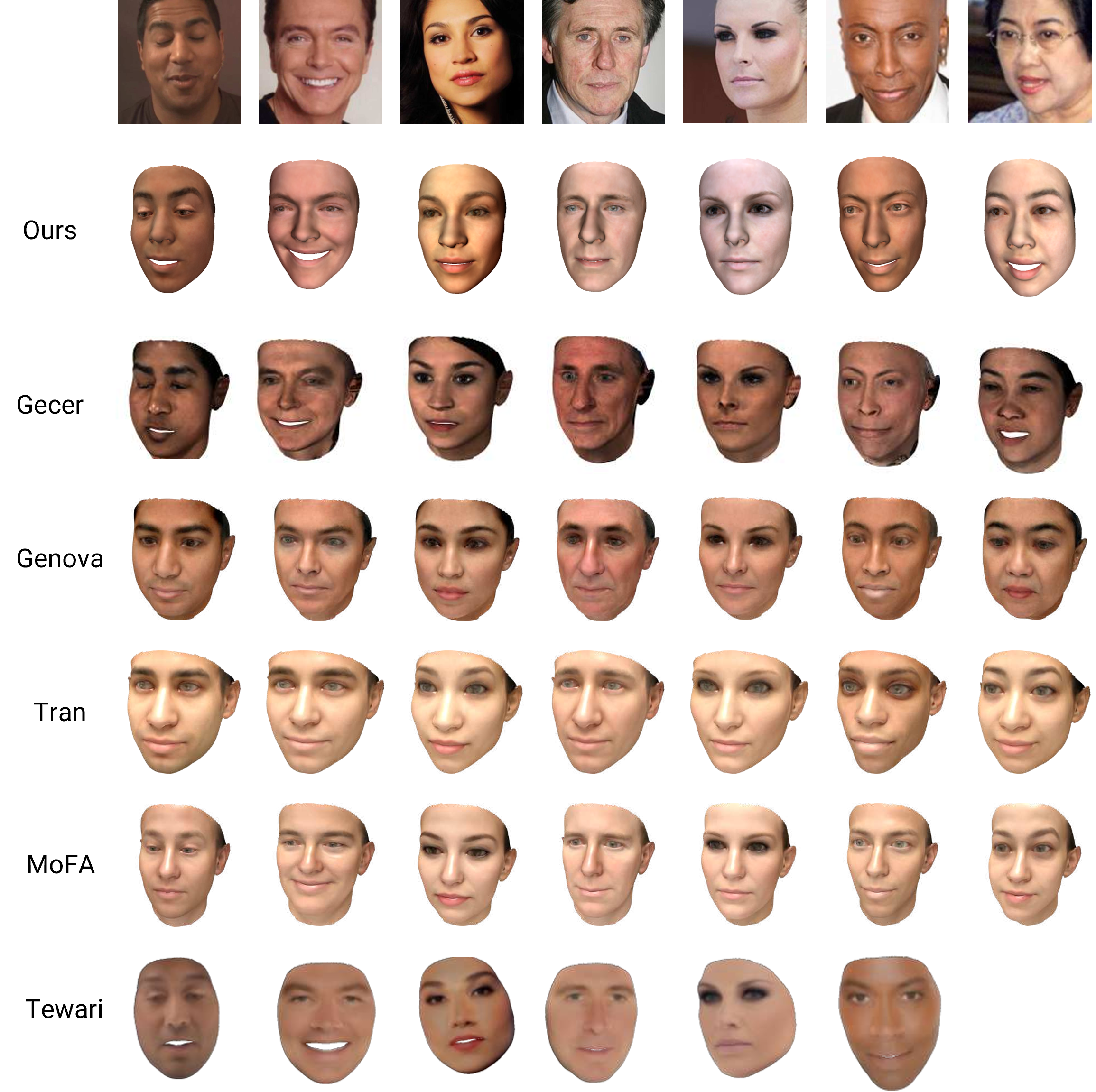}
  \caption[Qualitative comparison of rendered texture on MoFA-Test dataset]{Qualitative comparison of rendered texture results on the MoFA-Test dataset \cite{MoFA_Tewari_2017_ICCV}. Results from previous methods are taken from \cite{tfmeshrender_2018_CVPR, Gecer_2019_CVPR}, including Sela \textit{et al.} \cite{sela2017unrestricted}, Tewari \textit{et al.} \cite{tewari2018self}, MoFA \cite{MoFA_Tewari_2017_ICCV}, Tran \textit{et al.} \cite{Tran_2017_CVPR}, Genova \textit{et al.} \cite{tfmeshrender_2018_CVPR}, and Gecer \textit{et al.} \cite{Gecer_2019_CVPR}.}
  \label{fig:mofa_texture_compar}
\end{figure}

For the MoFA test dataset \cite{MoFA_Tewari_2017_ICCV}, as shown in Figure \ref{fig:mofa_shape_compar} and Figure \ref{fig:mofa_texture_compar}, our method demonstrates superior performance in both shape and texture reconstruction across a variety of samples. Also, the texture and shape reconstructions preserve the identity characteristics better than in other baseline methods. For shape reconstruction, our method excels in preserving facial geometry and capturing subtle details. For example, in the second input of Figure \ref{fig:mofa_shape_compar}, the open mouth is not accurately reconstructed by other methods, while MSMA achieves a realistic representation of the open mouth and accurately preserves the overall facial structure. Similarly, in the fifth input of Figure \ref{fig:mofa_shape_compar} with challenging head poses, our method outperforms others by maintaining symmetry and capturing consistent depth, whereas baseline methods exhibit distorted geometries or fail to represent occluded regions effectively. For the sixth input image of Figure \ref{fig:mofa_shape_compar}, MSMA preserves the geometry of the cheekbones and jawline better than competing methods. In terms of texture reconstruction, MSMA shows a remarkable ability to preserve identity-related details and produce realistic textures. For instance, in the fourth input of Figure \ref{fig:mofa_texture_compar}, baseline methods like Gecer \textit{et al.} fail to recover natural skin tones, resulting in artifacts and oversaturated regions. In contrast, our method achieves a smoother and more realistic texture representation. Similarly, for the seventh input of Figure \ref{fig:mofa_texture_compar}, MSMA produces textures that closely resemble the input identity, while other methods blur facial attributes and fail to reconstruct finer details such as wrinkles and skin patterns.

\begin{figure}[t]
  \centering
  \hspace{-1cm} % Adjust the value here to shift left
  \includegraphics[width=0.95\textwidth, height=0.5\textheight]{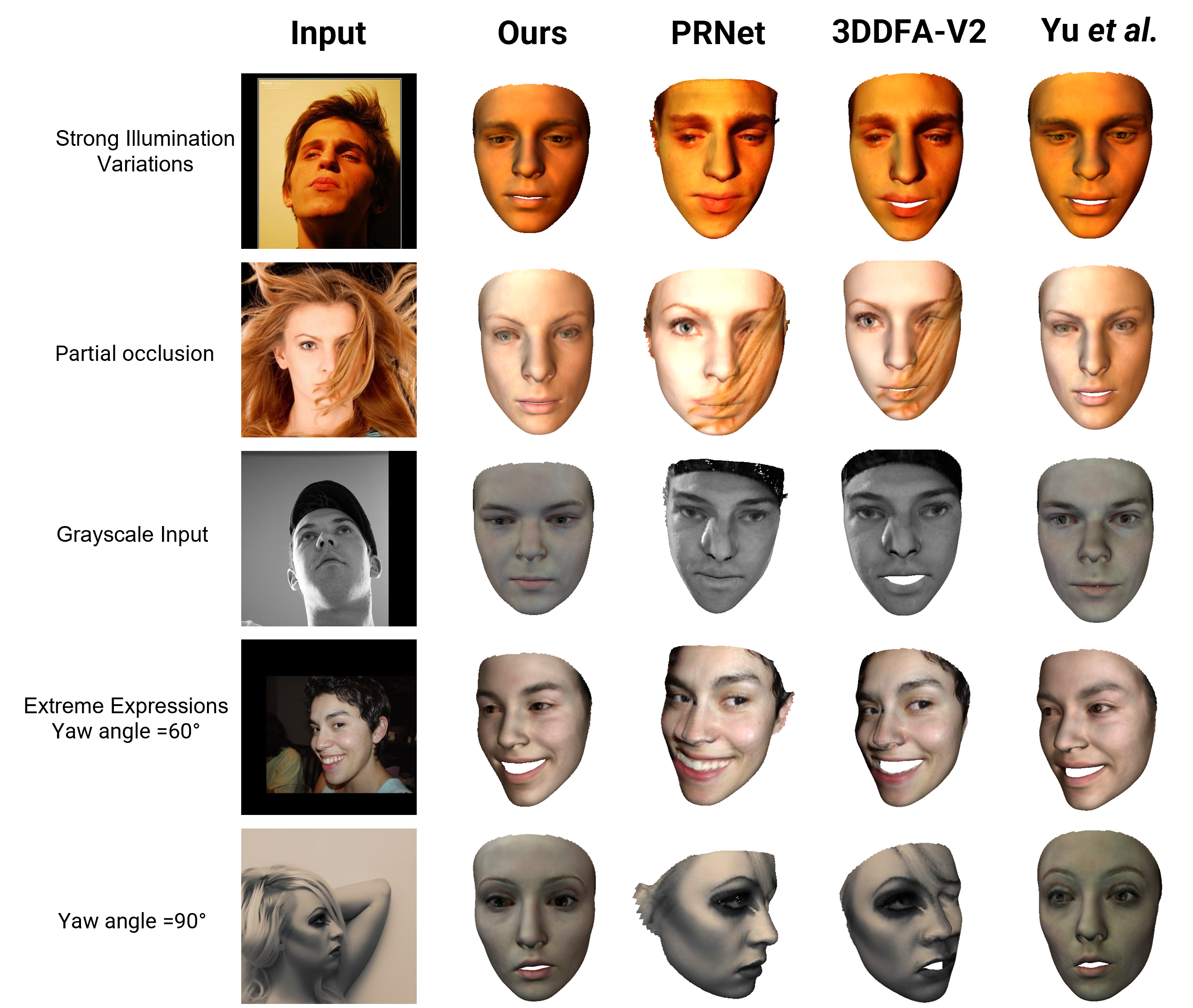}
  \caption[Qualitative comparison of 3D face reconstruction results across different methods on the AFLW2000-3D dataset.]{Qualitative comparison of 3D face reconstruction results across different methods on the AFLW2000-3D dataset \cite{ALFW-3D}. From left to right: input image, reconstruction results from our proposed MSMA framework, PRNet \cite{feng2018prn}, 3DDFA-V2 \cite{3DDFAv2}, and Yu \textit{et al.} \cite{deep3d_deng}. The rows represent challenging scenarios, including (1) extreme illumination variations, (2) occlusions (e.g., hair obstruction), (3) grayscale input images, (4) extreme facial expressions, (5) large pose variations, and (6) artistic/stylized facial representations. All input images are sourced from the AFLW2000-3D dataset.}
  \label{fig:AFLW_compar}
\end{figure}

For the qualitative comparison on the AFLW2000-3D dataset \cite{ALFW-3D}, as shown in Figure \ref{fig:AFLW_compar}, we compared our MSMA framework with PRNet \cite{feng2018prn}, 3DDFA-V2 \cite{3DDFAv2}, and Yu \textit{et al.} \cite{deep3d_deng}. The results highlight the superior performance of MSMA in reconstructing texture and geometry under diverse challenging conditions, including extreme lighting, occlusions, grayscale inputs, and large pose variations. For the first input, our method accurately reconstructs the mouth region, maintaining alignment and natural shape, even for subtle details of the slightly open mouth. In contrast, baseline methods produce simplified and overly smoothed results, failing to capture the fine details. For the second input, which includes hair occlusion and a partially closed mouth, MSMA successfully reconstructs the occluded region while maintaining realistic geometry for the mouth. Other methods either fail to handle the occlusion effectively or produce incomplete reconstructions, often introducing artifacts around the mouth. The third input further demonstrates the advantage of MSMA in preserving identity-related details. Fine structures like the nasal bridge and nasolabial folds are faithfully reconstructed, whereas competing methods, such as Yu \textit{et al.} \cite{deep3d_deng}, blur these features or fail to represent them distinctly. For the fourth input, which includes pronounced facial expressions, MSMA achieves a precise balance of facial structure and expressions, preserving critical features like the eyes and mouth without distortion. Other methods struggle to maintain structural consistency, often leading to exaggerated or misaligned regions. Finally, for the fifth input featuring significant lighting variations, MSMA showcases robust performance by accurately reconstructing both illuminated and shadowed facial regions. It successfully captures consistent lighting effects while maintaining facial symmetry. Baseline methods, however, fail to address lighting sensitivity, resulting in uneven or distorted reconstructions.
\begin{figure}[t]
  \centering
  \includegraphics[width=1.0\textwidth, height=0.5\textheight]{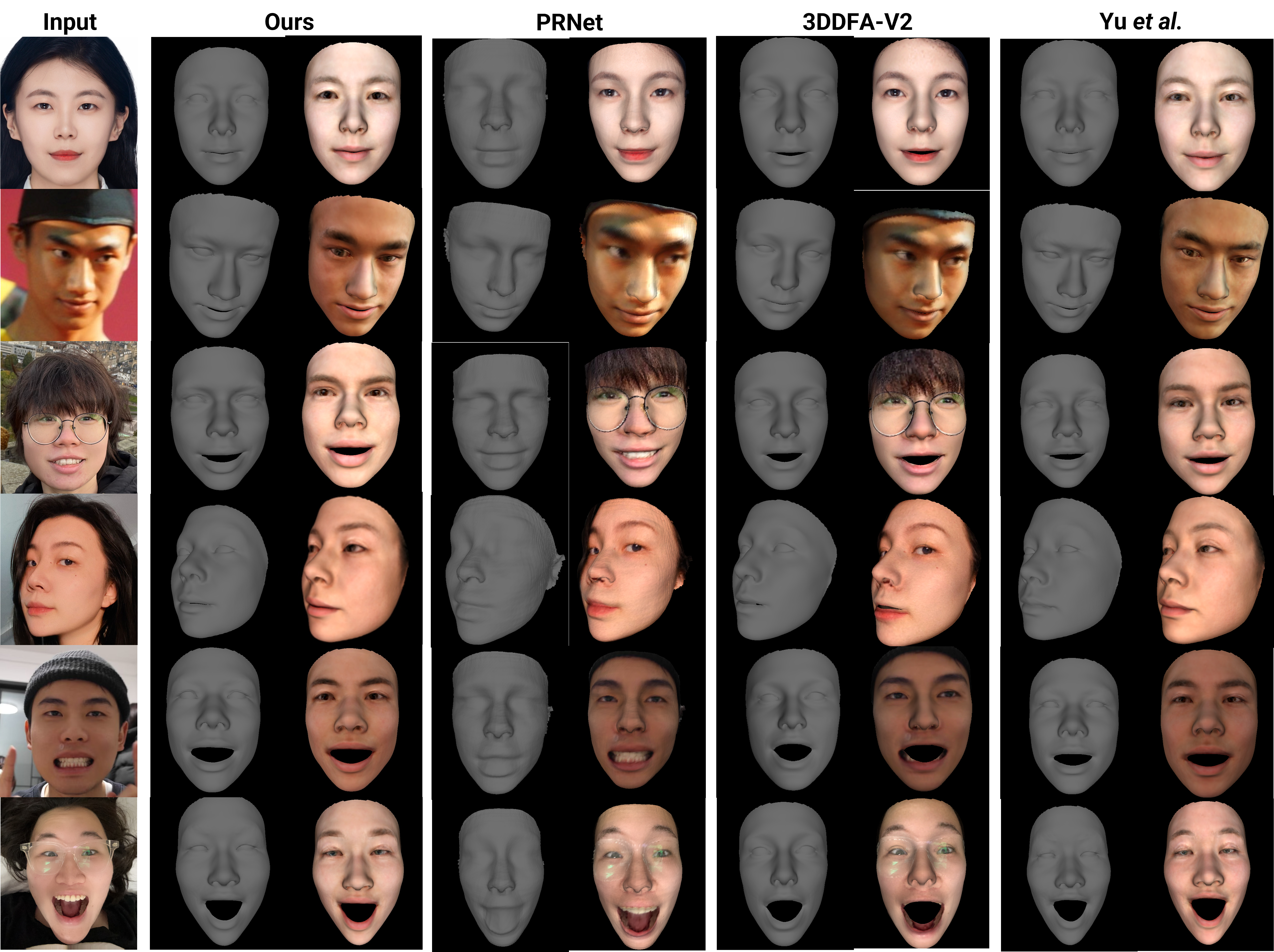}
\caption[Qualitative results of 3D face reconstruction on a custom-collected dataset]{Qualitative comparison of 3D face reconstruction results on a custom-collected dataset. From left to right: input image, reconstruction results from our proposed MSMA framework, PRNet \cite{feng2018prn}, 3DDFA-V2 \cite{3DDFAv2}, and Yu \textit{et al.} \cite{deep3d_deng}. Rows represent different conditions: (1) frontal view, (2) extreme illumination variations, (3) occlusions (e.g., glasses, bangs), (4) large pose variations, (5) extreme facial expressions, and (6) extreme facial expressions combined with occlusions (e.g., glasses coverage).}
  \label{fig:msma_custom_compar}
\end{figure} 

For the qualitative comparison on our custom-collected dataset, as shown in Figure \ref{fig:msma_custom_compar}, the proposed MSMA framework demonstrates superior 3D face reconstruction performance across a variety of challenging conditions. Specifically, for the first row with frontal views, MSMA accurately preserves facial geometry and identity attributes, outperforming baseline methods such as PRNet \cite{feng2018prn} and 3DDFA-V2 \cite{3DDFAv2}, which tend to produce over-smoothed or distorted features. For the second row, under extreme illumination variations, MSMA effectively captures natural skin tones and facial contours, while baseline methods struggle with exaggerated artifacts or uneven lighting effects. In the third row, which involves occlusions like glasses and bangs, MSMA robustly reconstructs occluded facial regions with minimal loss of detail, as opposed to other methods, which often exhibit incomplete reconstruction or visual artifacts around occlusion boundaries. The fourth row, highlighting large pose variations, shows that MSMA maintains geometric consistency and identity preservation across rotated angles, whereas PRNet and 3DDFA-V2 produce asymmetric or blurred outputs. For the fifth row, under extreme facial expressions, MSMA excels in capturing subtle deformations and facial dynamics, ensuring that features like the eyes and mouth remain coherent and realistic. In contrast, other methods often fail to handle these complex expressions, leading to distorted or flattened regions. Finally, in the sixth row, combining extreme expressions with occlusions, MSMA effectively balances facial structure and occlusion handling, accurately reconstructing key features while maintaining high identity fidelity. Baseline methods, however, fail to adequately address both challenges, resulting in noticeable inconsistencies or missing details.
\begin{table}[t]
\centering
\setlength{\tabcolsep}{10pt} 
\renewcommand{\arraystretch}{1.2}
\caption[Comparison of Mean Root Mean Squared Error (RMSE) for Face Reconstruction on the MICC Florence Dataset]{Comparison of Mean Root Mean Squared Error (RMSE) for Face Reconstruction on the MICC Florence Dataset (in mm). Results are shown as Mean $\pm$ Standard Deviation.} \label{table:MSMA_RMSE_MICC}
\begin{tabular}{lccc}
\toprule
\multirow{2}{*}{\textbf{Method}} & \multicolumn{3}{c}{\textbf{MICC Florence}}\\
\cline{2-4}
& Cooperative & Indoor & Outdoor\\
\toprule
RingNet \cite{withoutsupervision2019} & $2.09 \pm 0.48$ & $2.13 \pm 0.46$ & $2.10 \pm 0.47$ \\
CPEM \cite{CPEM} & $2.08 \pm 0.59$ & 2.09 $\pm$ 0.54 & $2.02 \pm 0.54$ \\
Yu \textit{et al.} \cite{deep3d_deng} & $1.74 \pm 0.58$ & $1.75 \pm 0.49$ & $1.79 \pm 0.46$ \\
Yu \textit{et al.} (Pytorch) \cite{deep3d_deng} & $1.67 \pm 0.49$ & 1.68 $\pm$ 0.51 & $1.71 \pm 0.53$ \\
Tran \textit{et al.} \cite{Tran_2017_CVPR} & $1.97 \pm 0.49$ & $2.03 \pm 0.45$ & $1.93 \pm 0.49$ \\
MGCNet \textit{et al.} \cite{MGCNet} & $1.78 \pm 0.55$ & 1.78 $\pm$ 0.54 & $1.81 \pm 0.59$  \\
3DDFA-v2 \cite{3DDFAv2} & \underline{1.65 $\pm$ 0.56} & \underline{$1.66 \pm 0.50$} & 1.71 $\pm$ 0.66 \\
Booth \textit{et al.} \cite{booth20173d} & $1.82 \pm 0.29$ & 1.85 $\pm$ 0.22 & \textbf{1.63 $\pm$ 0.16}  \\
Ours (MSMA)  & \textbf{1.64 $\pm$ 0.48} & \textbf{1.64 $\pm$ 0.49} & \underline{1.65 $\pm$ 0.51}  \\
\bottomrule
\end{tabular}   
\end{table}

  \subsection{Quantitative Results}
  We evaluate the 3D face reconstruction performance using the point-to-plane Root Mean Square Error (RMSE). For comparison, we include results from RingNet \cite{withoutsupervision2019}, CPEM \cite{CPEM}, Yu \textit{et al.} \cite{deep3d_deng}, Tran \textit{et al.} \cite{Tran_2017_CVPR}, 3DDFA-V2 \cite{3DDFAv2}, MGCNet \cite{MGCNet}, and Booth \textit{et al.} \cite{booth20173d} on the MICC Florence dataset and include results from MS-SFN \cite{MS-SFN}, CPEM \cite{CPEM}, PRNet \cite{feng2018prn}, 3DDFAv2 \cite{3DDFAv2}, FML \cite{FML}, and Yu \textit{et al.} \cite{deep3d_deng} on the FaceWarehouse dataset.

MICC Florence dataset includes 53 subjects, each with a high-quality neutral face scan and three video clips captured under different environmental conditions (cooperative, indoor, and outdoor) of increasing difficulty. For doing comparison on this dataset, we compute the point-to-plane RMSE by averaging errors across frames within each scenario (cooperative, indoor, outdoor) and then averaging across all three scenarios. To ensure consistent error measurement, we align the reconstructed 3D faces to ground truth meshes cropped to a 95mm radius around the nose tip, as proposed in \cite{genova2018unsupervised, deep3d_deng, Gecer_2019_CVPR}. The alignment is performed using the Iterative Closest Point (ICP) algorithm \cite{ICP} with isotropic scaling to reduce discrepancies, after which we compute the point-to-plane distances between the aligned meshes following the methodology described in \cite{tfmeshrender_2018_CVPR}.

The results of evaluation on the MICC Florence dataset are presented in \ref{table:MSMA_RMSE_MICC}. The values for Booth \textit{et al.} are derived from the GANFIT study \cite{Gecer_2019_CVPR}, while the results for RingNet and CPEM are sourced from the CPEM paper \cite{CPEM}. The performance of MGCNet is obtained from Qixin \textit{et al.} \cite{MM}, and the results for Tran \textit{et al.} \cite{Tran_2017_CVPR} are taken from Yu \textit{et al.} \cite{deep3d_deng}. The performance of Yu \textit{et al.} \cite{deep3d_deng}, Yu \textit{et al.} (PyTorch) \cite{deep3d_deng}, and 3DDFA-v2 \cite{3DDFAv2} were computed by running their pre-trained models using the methodology described above. From the result, it can be seen that our method achieves competitive reconstruction accuracy, with smaller RMSE values compared to previous works in most conditions. Our method outperforms other methods except for Booth \textit{et al.} in the outdoor setting on the Florence dataset.However, it is important to note that Booth \textit{et al.} utilized annotated data for supervised learning of the 3DMM and an optimization-based fitting procedure, which provides an advantage in specific scenarios but requires a greater reliance on labeled datasets.

The FaceWarehouse dataset includes 150 individuals, each captured with 20 different facial expressions, including neutral, mouth-opening, smiling, and kissing, among others. For comparison, we follow \cite{deep3d_deng} and use 9 subjects selected by \cite{tewari2018self}, resulting in a total of 180 images (9 subjects × 20 expressions). For this dataset, we calculate the point-to-point RMSE, which directly measures the mesh of the reconstructed faces relative to the ground truth.
\begin{table}[t]
  \centering
  \caption[Geometric reconstruction error (mm) on Facewarehouse (FW) dataset.]{Geometric reconstruction error (mm) on Facewarehouse (FW) dataset. Mean and standard deviation are displayed in separate columns.}
  \label{table:FW_error}
\begin{tabular}{lcccccccc}
    \toprule
     & \cite{MS-SFN} & \cite{CPEM} & \cite{feng2018prn} & \cite{3DDFAv2} & \cite{FML} & \cite{deep3d_deng} & \cite{deep3d_deng}-P & ours \\
    \midrule
    \textbf{Mean (mm)} & 2.20 & 2.03 & 2.20 & 2.12 & 1.90 & 1.94 & 1.79 & \textbf{1.77} \\
    \textbf{Std (mm)} & 0.45 & 0.33 & 0.45 & 0.49 & 0.40 & 0.50 & 0.49 & 0.46 \\
    \bottomrule
\end{tabular}
\end{table}

Following the methodology and evaluation protocol described in \cite{deep3d_deng}, we compute the point-to-point distances after alignment with ICP algoreithm between the reconstructed meshes and the ground-truth on samples from the Facewarehouse datahouse. The results are presented in \ref{table:FW_error}. The values for MS-SFN \cite{MS-SFN} and CPEM \cite{CPEM} are derived from the CPEM paper \cite{CPEM}, while the results for FML \cite{FML} are sourced from the its paper. The performance of PRNet \cite{feng2018prn}, Yu \textit{et al.} \cite{deep3d_deng}, Yu \textit{et al.} (PyTorch) \cite{deep3d_deng}, and 3DDFA-v2 \cite{3DDFAv2} were computed by running their pre-trained models using the methodology described above. From the results, it can be seen that our method achieves superior reconstruction accuracy, as reflected by the lowest mean error among the compared methods under diverse expression conditions.
\begin{figure}[t]
  \centering
  \includegraphics[width=1.0\textwidth]{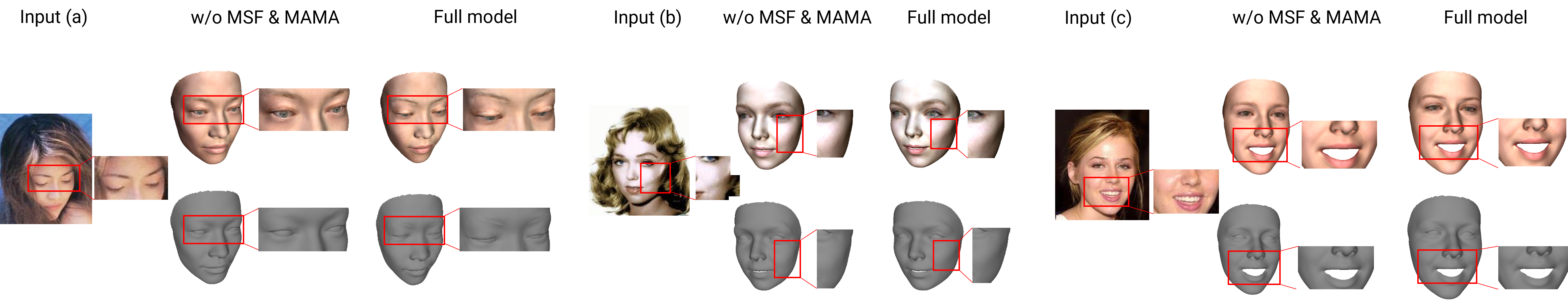}
\caption[Ablation Study on CelebA Dataset]{Ablation study on CelebA dataset between our full model and the model without modules. input (a): example with occulution. input (b): example with uneven illumination. input (c): example with low-light condition.}
  \label{fig:ablation_celeba}
\end{figure} 
  \subsection{Ablation Study}
In the ablation study, we have validated the contributions of the Multi-Scale Feature Fusion (MSF) module and the multi-attribute learning strategy in our MSMA framework, we conducted ablation experiments on the CelebA dataset. Specifically, we compared the performance of our full model with a baseline version that excludes both the MSF module and the multi-attribute learning strategy. Results for three challenging scenarios—(a) occlusion, (b) uneven illumination, and (c) smiling expression—are shown in \ref{fig:ablation_celeba}.
For the input with occluded eyes, the baseline reconstruction inaccurately renders the eyelids open, failing to adapt to occlusion. In contrast, the full model produces a natural representation of the closed eyelid, demonstrating the MSF module's ability to recover occluded regions and the MAMA strategy's role in optimizing facial attributes like eyelid texture.
For the input under uneven illumination, the baseline output exhibits oversmoothed cheek textures and diminished contour details. The full model accurately reconstructs finer cheek details and texture variations, reflecting the effectiveness of the MSF module in capturing global and local features and the MAMA strategy in adapting to environmental lighting changes.
For the smiling expression, the baseline struggles to reproduce pronounced nasolabial folds, resulting in a flatter reconstruction. The full model captures these folds with greater depth and detail, ensuring a more realistic representation of the expression. This highlights the importance of multi-scale fusion and attribute-specific optimization in preserving both fine details and structural coherence.
   
  \section{Conclusions}
In this paper, we proposed a Multi-Scale Feature Fusion with Multi-Attribute (MSMA) framework to advance the accuracy and fidelity of 3D face reconstruction from single unconstrained images. The core of MSMA lies in its Multi-Scale Feature Fusion (MSF) module, which aligns and integrates multi-resolution feature maps extracted from shared backbone features. By leveraging adaptive upsampling and downsampling strategies, the MSF module ensures precise alignment of hierarchical features across scales, enabling the effective disentanglement and optimization of task-specific facial attributes, including geometry, texture, pose, and illumination. This multi-scale fusion process captures complementary global and local facial characteristics, enhancing both structural coherence and fine-grained detail recovery.

Our training process utilizes a weakly-supervised strategy, combining 3D Morphable Model (3DMM) parameters with a differentiable renderer to enable end-to-end learning without requiring ground-truth 3D face scans. Extensive experiments conducted on benchmark datasets, including MICC Florence, FaceWarehouse, and a custom-collected dataset, demonstrate the robustness and adaptability of the MSMA framework. The proposed method achieves reliable and high-quality results in image-based geometry and texture reconstruction, validated through both quantitative and qualitative evaluations, and outperforms or remains competitive with ten existing state-of-the-art methods. Furthermore, ablation studies underscore the effectiveness of the Multi-Scale Feature Fusion (MSF) module and multi-attribute learning strategy, which ensures that each task-specific attribute is derived from the most relevant scale of feature representation. This integration of multi-scale fusion with attribute-specific optimization establishes the robustness and precision of MSMA, offering a novel perspective on advancing single-view 3D face reconstruction. 

\section*{Acknowledgments}
This work was completed during the Danling's MPhil studies at the University of Manchester. Thanks to Professor Iacopo Masi and Professor Stefano Berretti, for supporting MICC Florence dataset at the University of Florence, as well as to Professor Kun Zhou for supporting Facewarehouse dataset at Zhejiang University. Thanks to the Danling's friends who allowed the author to collect and test their personal facial images.
%Bibliography
\bibliographystyle{unsrt}  
%\bibliography{references}  
\bibliography{paper}

\end{document}